\documentclass[conference]{IEEEtran}
\IEEEoverridecommandlockouts
\usepackage{cite}
\usepackage{amsmath,amssymb,amsfonts}
\usepackage{algorithmic}
\usepackage{graphicx}
\usepackage{textcomp}
\usepackage{xcolor}
\usepackage{array}
\usepackage{colortbl}
\usepackage{diagbox}
\usepackage{footnote}
\usepackage{hyperref}
\usepackage{tabularx}
\usepackage{booktabs}
\usepackage{float}

\usepackage{lipsum}

\newcommand\blfootnote[1]{%
  \begingroup
  \renewcommand\thefootnote{}\footnote{#1}%
  \addtocounter{footnote}{-1}%
  \endgroup
}

\def\BibTeX{{\rm B\kern-.05em{\sc i\kern-.025em b}\kern-.08em
    T\kern-.1667em\lower.7ex\hbox{E}\kern-.125emX}}

\makeatletter
\newcommand{\thickhline}{%
	\noalign {\ifnum 0=`}\fi \hrule height 2pt
	\futurelet \reserved@a \@xhline
}
\newcolumntype{"}{@{\hskip\tabcolsep\vrule width 2pt\hskip\tabcolsep}}
\makeatother

\begin{document}

\title{Many Heads but One Brain: Fusion Brain -- a Competition and a Single Multimodal Multitask Architecture}

\author{
\IEEEauthorblockN{Daria~Bakshandaeva$^{*1}$, Denis~Dimitrov$^{*2}$, Vladimir~Arkhipkin$^2$, Alex~Shonenkov$^2$, Mark~Potanin$^2$, \\ Denis~Karachev$^2$, Andrey Kuznetsov $^2$, Anton Voronov$^2$, Vera Davydova$^2$, Elena Tutubalina$^2$, \\
	Aleksandr~Petiushko$^2$}
\IEEEauthorblockA{
$^1$ University of Helsinki, $^2$ Artificial Intelligence Research Institute\\
daria.bakshandaeva@helsinki.fi, \{ext-dimitrov,AVShonenkov,potanin.m.st,ext-kuznetsov,voronov,Petiushko\}@airi.net, \\ veranchos@gmail.com, tutubalinaev@gmail.com, arkhipkin.v98@gmail.com, denis.karachev@ocrv.ru
}
}

\maketitle

\begin{abstract}
Supporting the current trend in the AI community, we present the AI Journey 2021 Challenge called Fusion Brain, the first competition which is targeted to make the universal architecture which could process different modalities (in this case, images, texts, and code) and solve multiple tasks for vision and language. The Fusion Brain Challenge combines the following specific tasks: Code2code Translation, Handwritten Text recognition, Zero-shot Object Detection, and Visual Question Answering. We have created datasets for each task to test the participants' submissions on it. Moreover, we have collected and made publicly available a new handwritten dataset in both English and Russian, which consists of 94,128 pairs of images and texts. We also propose a multimodal and multitask architecture -- a baseline solution, in the center of which is a frozen foundation model and which has been trained in Fusion mode along with Single-task mode. The proposed Fusion approach proves to be competitive and more energy-efficient compared to the task-specific one. 
\end{abstract}


\section{Introduction}

A significant part of the information perceived by a person and required for making even the simplest everyday decisions is presented in multiple modalities, that is, with the help of different types of ``input information'', requiring the use of various senses and types of knowledge. Visual information requires visual perception, processing natural language texts presupposes the knowledge of the language, auditory information implies the perception and analysis of sound, and so on. Each of these modalities is handled by separate, sometimes overlapping areas of machine learning and artificial intelligence: computer vision, natural language processing, speech processing, video processing, etc. \blfootnote{*Both authors contributed equally to this research.}

However, a successful solution to emerging problems often cannot be obtained by analyzing data coming from only one modality, just as it is not always sufficient for a human being to use only sight or only hearing to make a rational decision. In such cases, information required to solve such problems can be divided into several ``input types'', called data modalities, all of which should be taken into consideration to make successful decisions.

Multi-task learning has a long history mostly in the natural language processing domain. One of the possible reasons is that having the correct representation and thus ``understanding'' of text passage, one can solve many downstream tasks: sentiment analysis, question answering, language translation etc. One of the most widely used approaches here is to have the lower (encoding) layers shared for all tasks, while having the upper layers (also called ``heads'') task-specific and learned separately \cite{liu2019multi}.

It is only recently that scientists have proposed to combine multi-modality and multi-task in one model, taking the joint approach: using different encoders for different modalities, then combining different types of information during middle processing, and completing the process with task-specific heads - e.g. the UniT \cite{hu2021unit} approach, where visual and textual modalities are used, and 7 tasks of computer vision (e.g. object detection), text processing (e.g. sentiment analysis) and vision-and-language (e.g. visual question answering) fields are solved. 

\begin{figure}[t!]
    \centering
    \includegraphics[width=0.5\textwidth]{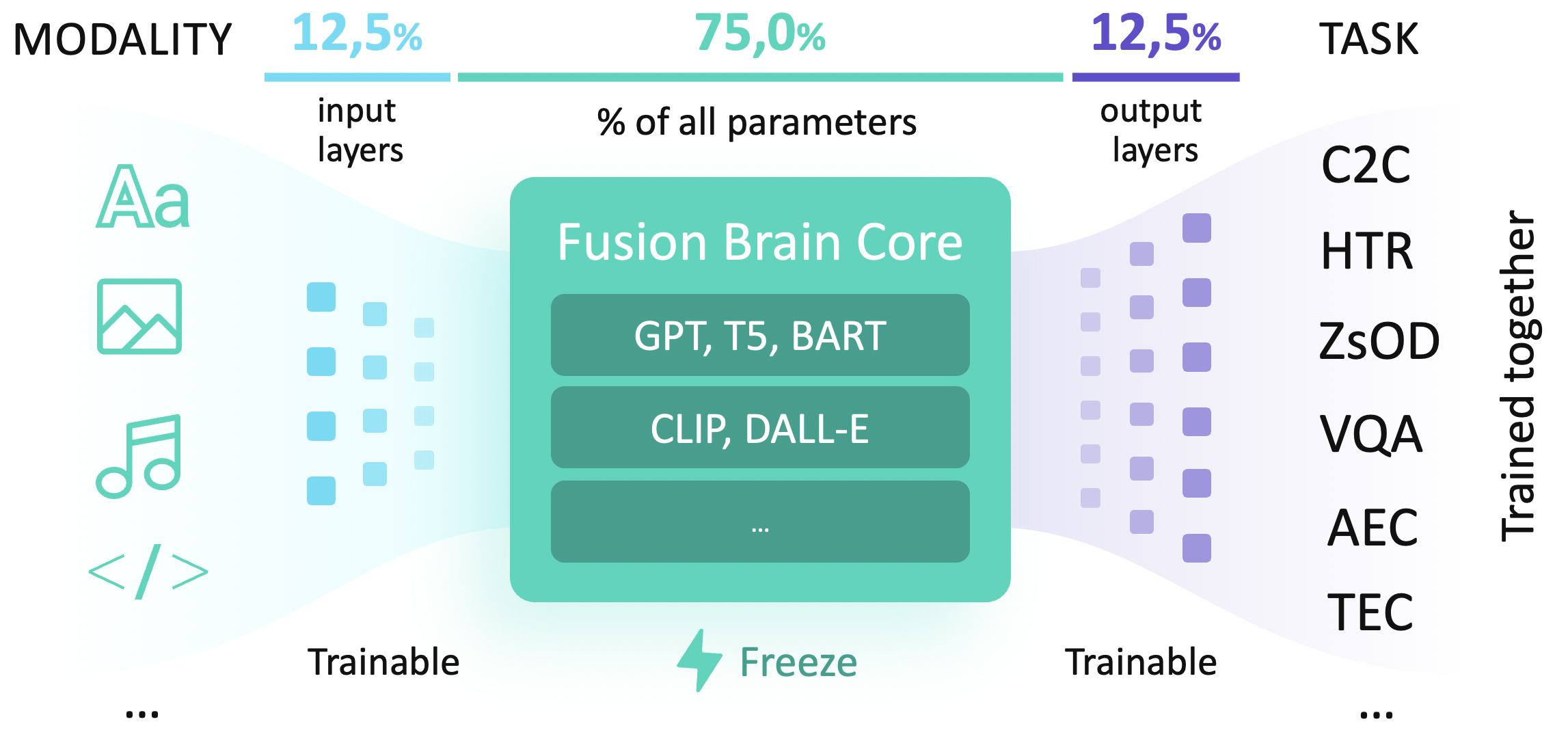}
    \caption{Concept of the multi-modal and multi-task architecture Fusion Brain. The tasks here are C2C -- Code2code Translation, HTR -- Handwritten Text Recognition, ZsOD - Zero-shot Object Detection, VQA - Visual Question Answering, AEC - Audio Emotion Classification, and TEC - Text Emotion Classification}
    \label{fig:conceptfusionbrain}
\end{figure}

The problem of training large pretrained multi-modal and multi-task models can be separated into 2 subtasks: 1) How to combine modalities, and 2) How to combine tasks.

As for the first question, the current state-of-the-art research in the multi-modal processing is mostly focusing on the questions of the stage at which modalities should be fused (``early'', ``middle'' or ``late'' fusion) and the ways to implement this fusion (through iterative processing or by a modality bottleneck) \cite{liang2018multimodal,li2019visualbert,das2020detecting,savchenko2020ad}. The important approaches for modality fusion are Perceiver \cite{jaegle2021perceiver} and Perceiver IO \cite{jaegle2021perceiverio}, where the modality-specific information serves as the key-value for iterative cross-attention and is later processed by GPT-2-like \cite{radford2019language} transformer. Cross-attention blocks are also used in Flamingo \cite{flamingo} to incorporate the information from the pretrained visual encoder into the frozen language model (Chinchilla \cite{chinchilla}), thus allowing to process multimodal (visual and textual) inputs. 

Another interesting and promising example of sharing the modality information is the so-called multimodal bottleneck transformer (MBT) \cite{nagrani2021attention}, where the fusion of the modalities is done: a) closely to the top of the transformer layers; b) only through a very small number $B$ of multimodal neurons (in the work $B = 4$ is used) making the cross-modality sharing only through a small bottleneck, which proves to be very efficient. Finally, incorporation of different modalities (like RGB and OpticalFlow) inside the single model via mutual modality learning  can be used \cite{komkov2020mutual}.

The combination of tasks can also be implemented in different ways. An approach similar to above-mentioned UniT is the so-called frozen pretrain transformer (FPT) technique \cite{lu2021pretrained}, which is a source of inspiration for our proposed baseline. However, such multitask pipeline, when different tasks/modalities are processed through separate heads, is not the only one. The more interesting approaches use the more sophisticated ways of dealing with multiple tasks: for instance, they incorporate either the task-specific adapters \cite{houlsby2019parameter,pfeiffer2020adapterfusion} between the frozen layers or the fully learnable (trainable) task representation (embedding) that can be later propagated in the non-trivial way through the major part of the model (see Perceiver IO, HyperGrid \cite{tay2020hypergrid} or conditionally adapted approach \cite{pilault2020conditionally}).

As different tasks within the domain can have similar formats, in general-purpose agent Gato \cite{gato}, a transformer decoder model, it is proposed to use prompt conditioning instead of simple one-hot identifiers in order to disambiguate tasks: while training, for 25\% of the sequences in the batch, a prompt sequence generated by the same agent on the same task, is added. In half of these cases, the sequence is taken from the end of the episode (goal conditioning is obtained), in the other half, it is randomly sampled from the episode. Using prompt conditioning, as the authors note, would be ideal for adjusting to new tasks, if not the model's maximum context length restraint which doesn't allow the agent to get an access to the information sufficient to solve the desired problems. 

The corresponding research in the field of information retrieval (IR) is also worth mentioning. For now, however, it seems that quite straightforward solutions are used for IR, e.g. the combination of all task-specific datasets for training NLP model for multiple tasks \cite{maillard2021multi}, or the processing of multimodal data with the single transformer using the representations obtained by modality-specific encoders as the inputs for the multimodal retrieval \cite{gabeur2020multi, dzabraev2021mdmmt}.

We aim to promote the development of such promising and challenging field as multimodal and multitask research. Our main contributions are the following:
 \begin{itemize}
     \item prepared the data, task statement and leaderboard for the Fusion Brain Challenge;
     \item proposed the specialized as well as the overall metric to evaluate the models;
     \item created the simple yet efficient baseline which combines multimodal as well as multitask approach.
 \end{itemize}

\section{Tasks}
\label{sec:tasks}

 Within the competition we proposed to solve 4 subtasks:
 \begin{enumerate}
     \item Code2code translation (C2C),
     \item Handwritten text recognition (HTR),    
     \item Zero-shot object detection (ZsOD), 
     \item Visual question answering (VQA).
 \end{enumerate}

All subtasks which include natural language data are bilingual -- contain samples in both English and Russian. In the following subsections we will discuss each of the subtasks in more details.

\subsection{Subtask 1 - Code2code Translation}

Among the various problems within ML4Code field, the task of translating code snippets from one programming language (PL) to another was chosen. Even though source code can be attributed to text modality, it is definitely more structured than natural language, thus we would like to distinguish between them. The proposed task not only adds ``code modality'' to the challenge but also imposes the requirement for the model to be multilingual since it has to understand and generate code in two PLs.

Our C2C task requires a model to translate code snippets from Java to Python. The choice of such a pair of PLs induces extra complexity to the problem since translation between statically- and dynamically-typed languages is more intricate than translation between PLs with the same type checking. 

For training we proposed to use a dataset presented in~\cite{avatar}. AVATAR is a parallel corpus that consists of solutions written in Java and Python for 8,506 programming problems collected from competitive programming sites, online platforms, and open source repositories. We used solutions of 6,807 tasks from AVATAR for train, leaving 1,699 examples for the public part of the test set. The private test dataset was designed as follows: at first, Python snippets with a length corresponding to that of the 90th percentile of AVATAR test set part written in Python (up to 282 tokens obtained after tokenization \cite{pytok}) were retrieved from CodeNet~\cite{codenet} dataset; these code snippets were translated to Java by three annotators and then cross-checked; at the final stage, Java functions (not longer than 356 tokens, which matches the 90th percentile of the public test requests' lengths) were back-translated to Python and cross-checked as well to ensure that Python snippets generate the same outputs as source functions when given the same inputs. The resulting number of Java-Python pairs is 322.

CodeBLEU~\cite{CodeBLEU} is selected as an evaluation metric for this task. 
%See chapter~\ref{sec:metrics} for more details.

\subsection{Subtask 2 - Handwritten Text Recognition}

Handwritten Text Recognition is the task that naturally combines image and text modalities; the model is given an image with a handwritten piece of text in Russian or English and is required to transcribe it into digital text as an output. The dataset for this task was manually collected and annotated; it is composed of the examples from school notebooks. The training data consist of 66,599 images of words written in Russian language (participants of the Challenge may use an open datasets with forms of handwritten English text, e.g., IAM Handwriting Database \cite{htrdb}). The public test set includes 14,973 images: 5,973 in English and 9,000 in Russian. The private test part consists of 12,556 images, 5,494 of which are in English and 7,062 -- in Russian.
In total, our new handwritten dataset contains 82,661 images of Russian words, which makes it the largest Russian handwritten dataset in the world so far. We have also released this dataset \cite{htrdatasets} for the benefit of the research community. 

The evaluation metric for this task is string accuracy - the proportion of cases in which the predicted text (string) coincides with the ground truth transcription.

%Metric of evaluation for this task is accuracy. 

\subsection{Subtask 3 - Zero-shot Object Detection}

ZsOD task sets the following problems to the model: firstly, the model should accurately predict bounding boxes for various objects depicted in the images, given the descriptions of these objects in natural language \cite{xiuye2021ZsOD}. In our case, such a common computer vision task as object detection is complicated by the fact that there is no set of predefined classes to choose from -- a model is expected to detect classes not present in the training set (i.e. in a zero-shot regime). During inference, a model receives image-query pairs; a query is formatted as a list of textual descriptions (in Russian or English) of objects to detect. The query may contain entities that are absent in the image; a model should predict an empty list as a bounding box for such objects. 

The public test dataset is formed from a part of the VisualGenome~\cite{visualgenome} dataset (1,000 examples); the set of classes in it was hidden from the participants. Region descriptions from VisualGenome are used as positive classes (descriptions are normalized: reduced to lowercase; non-printable characters are removed, etc.; boxes related to the same entity are combined under a single description); negative classes are formed by replacing some objects/attributes in the description with those that are missing in the photo. For example, ``a grey chair'' is replaced by ``a \textit{pink} chair''. Also, descriptions of objects belonging to the same domain as the correct classes are used as negative examples: if the photo shows a street, then as negative examples there may be, for instance, descriptions such as ``tall green bricks wall'', ``shingled home in distance'', ``food stand in the street'' (provided, of course, that the described objects are not in the photo). The images for the private test set were either extracted from YFCC100M dataset~\cite{yfcc} or crawled from the Internet. In total, 827 images were attributed with positive (the descriptions of objects which are present on the photo) and negative (the descriptions of missing objects) labels by 10 annotators. The number of positive classes varies from 7 to 10 -- the same held true for the negative ones. For a specific image, descriptions can be either in English or in Russian. There can be more than one bounding box for a particular description in the queries, a perfect model should predict all of them. 

The F1-score metric is used for evaluation. Refer to the section~\ref{sec:f1-zsod} for more details.

\subsection{Subtask 4 - Visual Question Answering}

VQA is a classical multi-modal task that requires model to understand a textual question and generate an answer to it based on the corresponding image. The peculiarity of the problem is that the questions are not homogeneous: a correct answer can either consist of several words, or be monosyllabic (a ``yes/no'' answer) or be a number. It is assumed that only one answer per question is required. As with other tasks, the model should be bilingual in order to perform well, since questions can be expressed in both English and Russian and the answer is expected to be in the same language except when the question concerns the text on the image. For example, when the question is ``What is written on the T-shirt?'' the answer should be in the same language in which the text is written.

The public test dataset consists of questions in both Russian and English: the Russian-language part is translated examples from the first 10 thousand samples of the validation part of the VQA v2 dataset, the English part - next 10 thousand original samples from the same dataset. The public test set size is 5,446 examples. The private test set was compiled similarly to the one for ZsOD task, except for the nature of annotation: for each image (in total, 1,000 images), 6 questions in Russian or English and corresponding answers were formulated, resulting in 6,000 samples. The intersection with the private test set for ZsOD task is 724 images.

The evaluation metric for this task is accuracy. Each question has a list of possible correct answers; if the prediction matches at least one of the ground truth answers, it is considered true positive. 

\section{Baseline}

\begin{figure*}
  \includegraphics[width=\textwidth]{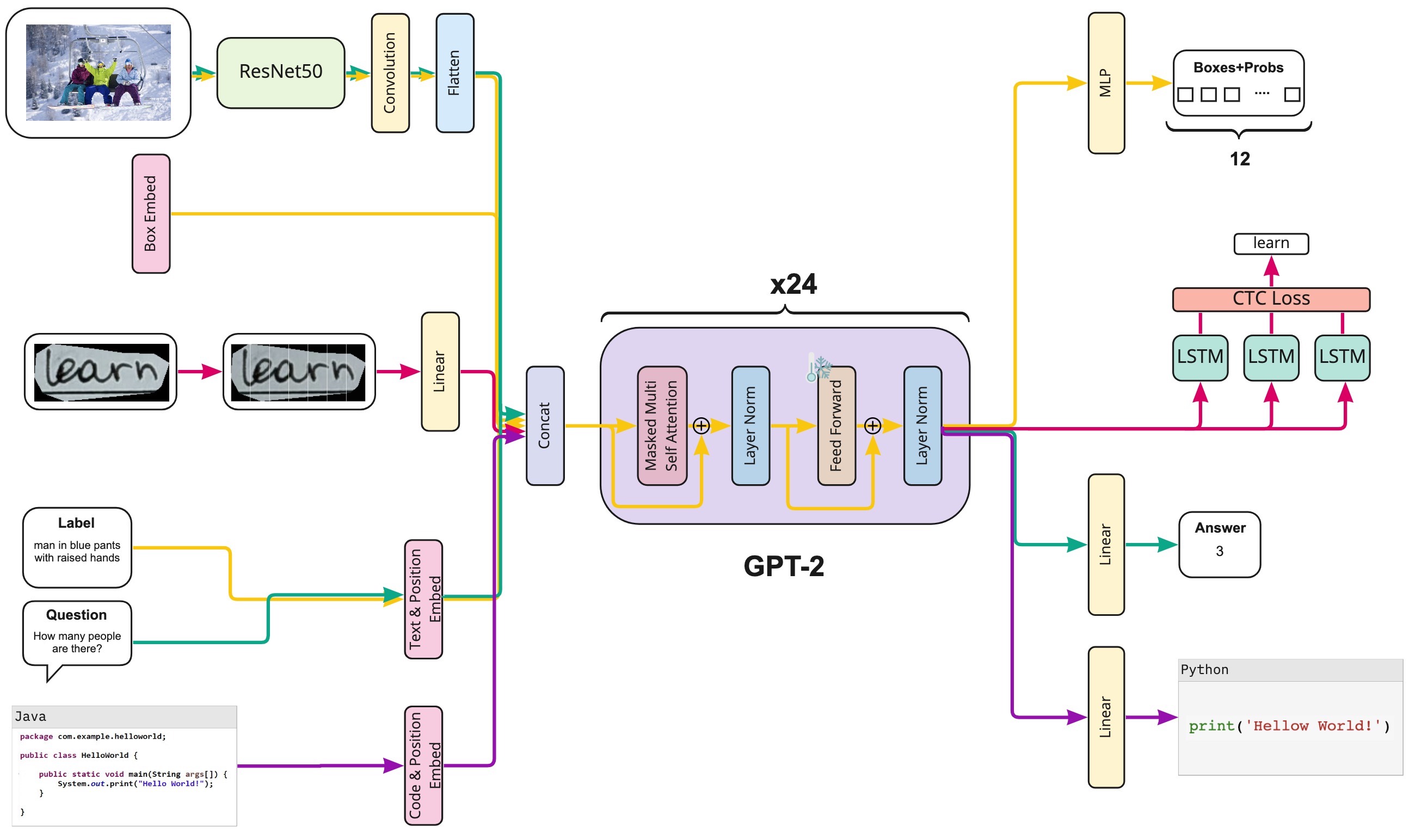}
  \caption{Baseline architecture}
  \label{baselinearch}
\end{figure*}

We provide a concept \cite{fbconcept} of a single model that is trained on several tasks related to different modalities (visual, audio and text). The concept is inspired by a work \cite{lu2021pretrained} that examines the ability of pretrained language models based on the Transformer architecture to form qualitative representations of arbitrary data sequences -- thus, generalizing to other modalities with minimal finetuning. The basis of the architecture proposed in the concept is the pretrained GPT-2 \cite{radford2019language} language model; experiments are carried out with a model which feed-forward layers are frozen.

% experiments are carried out both with a ``frozen'' model (with only output layer being finetuned), and with a model in which all layers are trained on three modalities simultaneously. 
We build our baseline solution also on top of Frozen Pretrained Transformer. The overall architecture can be seen on Figure \ref{baselinearch}. The core, the ``shared brain'' of the whole pipeline is GPT-2 Medium, pretrained on natural language; each type of data for a particular task undergoes its specific transformations in order to match the GPT-2's input format, and also has its specific head to generate predictions in accordance with the task. The input and output layers for each of the subtasks are described below.

It is worth mentioning that one can use any of the so-called foundation model (see, e.g., in-depth report \cite{foundationmodels}) instead of GPT-2 as Fusion Brain Core (see Figure \ref{fig:conceptfusionbrain}). Following the researchers from Stanford University CRFM we define foundation models as models trained on broad data at scale such that they can be adapted to a wide range of downstream tasks. Pretty nice examples of such models are BERT \cite{devlin2019bert}, BART \cite{lewis2019bart}, T5 \cite{raffel2020exploring}, GPT-3 \cite{brown2020language}, CLIP \cite{clip_open_ai}, DALL-E \cite{ramesh2021zeroshot}.

A research on Gato \cite{gato}, which became publicly available in 2022, when Fusion Brain Challenge had passed and a baseline model had been released, proves that such approach -- converting data of different modalities into flat sequence of tokens and then processing it with a single transformer decoder -- has great potential: a model with a single set of weights can solve 450 out of 604 tasks it was trained on, at over a 50\% expert score threshold.   

\subsection{C2C (code)}

As code is similar to natural language (although it is certainly more structured; the problem of choosing the best representation of source code goes beyond the scope of this work), no major transformations are needed in order to prepare the data for processing with GPT-2. The task is solved in decoder-only machine translation manner: during training, the source sequence (code snippet in Java) is concatenated with the target one (in Python) through the SEP token; the resulting sequence is fed into the GPT-2 with LM head on top in order to minimize the Categorical Cross-Entropy (CCE) loss \cite{Rubinstein99thecross-entropy}. When trained, the model auto-regressively generates Python code given Java function. 

\subsection{HTR (image)}

It is somewhat remarkable that images can also be processed using a language model and the proposed method. At first, raw images are subjected to smart resizing with proportions being preserved and empty space being padded; these resized images are then converted into vertical patches with full height and width equal to 8 pixels: $3 \times H_0 \times W_0 \rightarrow 3 \times 128 \times 512 \rightarrow 64 \times (128 \times 8 \times 3)$. Image patch features are extracted with a linear projection layer in order to match the size of the GPT-2 embedding space (1280) before being processed with GPT-2. The transformer outputs are then passed through LSTM and linear layers. The training process is based on the Connectionist Temporal Classification (CTC) loss \cite{ctc} that shows high performance in handwritten text recognition task \cite{shonenkov2021stackmix,de2019no,michael2019evaluating,DBLP:journals/corr/abs-2103-09354}. 

% At first, all input images are resized from $x_{img} \in \mathbf{R}^{3 \times H_0 \times W_0}$ to $x_{img} \in \mathbf{R}^{3 \times 128 \times 512}$. Reshaped images are then converted into a sequence of flattened patches $x_{p} \in \mathbf{R}^{64 \times (128 * 8 *3)}$ like in Vision Transformer. Vertical patches are used: the patch height is equal to the image height and the patch width is 8. Before being fed to GPT-2, each flattened element is passed through a linear projection layer in order to match the size of GPT-2 embedding space (768). Transformer outputs are then passed through LSTM and linear layers. The resulting matrix is of shape 64 × 152, where 152 is the number of unique characters and 64 is the number of patches. This matrix contains probabilities of possible characters for each position. The training process is based on the Connectionist Temporal Classification (CTC) loss \cite{ctc} which is widely used in text recognition task. 

\subsection{VQA and ZsOD (image + text)}

The proposed pipelines for solving VQA and ZsOD tasks are similar. Raw images are resized, processed with a convolutional backbone (three blocks of ResNet-50) \cite{he2015deep}; the resulting image feature map is then passed to Conv2D layer with a kernel size equal to $1$ and Flatten layer in order to match the size of the embedding space before processing with GPT-2 Medium:
$3 \times H_0 \times W_0 \rightarrow 3 \times 224 \times 224 \rightarrow  (14 \times 14) \times 1024 \rightarrow  196 \times 1024$. Texts are converted to tokens with the pretrained GPT-2 tokenizer, processed with token and position embeddings. The text format for VQA task is the following: \textit{``Question: '' + question + `` Answer: '' + answer + ``.''}; for ZsOD task: \textit{``Request: '' + text + ``.''}. The image and text embeddings are concatenated into one sequence: in case of VQA, text embedding follows image embedding; in case of ZsOD, it is vice versa. 

For VQA, the transformer outputs corresponding to text embeddings are passed through the linear layer in order to get a projection which is consistent with the dimension of the vocabulary. Cross-Entropy loss is used when adjusting model weights during training -- and only for the text tokens. The answer is generated auto-regressively. 

For ZsOD, 12 trainable box embeddings are introduced and concatenated with the image and text embeddings before being fed into the transformer. The output of GPT-2 is passed through MLP -- the resulting dimension is $12\times 5$: for each of 12 boxes, 4 coordinates and a probability score are obtained. The loss function used is similar to the one introduced in \cite{mdetr}: given \textit{M} ground-truth boxes and \textit{N} predicted boxes ($N\geq M$), \textit{M} predicted boxes are chosen so that 
\begin{eqnarray}
IoU(gt\_boxes[:, :4], pred\_boxes[: ,:4]) \\ 
- pred\_boxes[:, -1] \\ 
+ L1(gt\_boxes[:, :4], pred\_boxes[: ,:4])
\end{eqnarray}
is minimal.  

For the selected boxes, GIoU and L1 losses are minimized; for probability score, Binary Cross-Entropy Loss is used (1 is assigned to the selected boxes, 0 -- to the rest).

% The transformer outputs (both for text tokens and image feature map) are then projected with a linear layer to a shared semantic space using InfoNCE loss \cite{oord2019representation} like in CLIP \cite{clip_open_ai}. 

% In case of VQA, text projections are used as queries (Q), image feature map projections are used as keys (K) and values (V). The output of MMCA blocks is passed through the linear layer in order to get a projection corresponding to the dimension of the vocabulary. CCE loss is used when adjusting model weights during training. The answer is generated auto-regressively.

% In case of ZsOD it is vice versa: image feature map projections are used as Q, text projections are used as K and V. The output of MMCA blocks is passed through the adaptive max pool layer to reduce the amount of resulting bounding boxes per text query to 8 items. The bounding box predictions in the format of (x, y, w, h, probability score) are generated using MLP layers with Binary Cross-Entropy (BCE) loss \cite{Rubinstein99thecross-entropy}, Generalized Intersection over Union \cite{Rezatofighi_2018_CVPR} loss and L1 loss \cite{mae_article}.

\section{Experiments}

% \renewcommand{\arraystretch}{2}
% \begin{table}[h]
%   \label{tab:scores}
%   \begin{center}
%     \begin{tabular}{ |c|c|c|c|c| }
%     \hline
%     %\toprule
%   \shortstack{\textbf{Training} \\ \textbf{setup}} &  \shortstack{\textbf{C2C} \\ \textbf{CodeBLEU}} &  \shortstack{\textbf{HTR} \\ \textbf{Acc}} & \shortstack{\textbf{ZsOD} \\ \textbf{F1}} & \shortstack{\textbf{VQA} \\ \textbf{Acc}} \\
%     %\midrule
%     \hline
%     Single-task  & 2.11/2.11 & 2.11/2.11 & 2.11/2.11 & 2.11/2.11 \\
%     \hline
%     Fusion (HTR init.) & 2.11/2.11 & 2.11/2.11 & 2.11/2.11 & 2.11/2.11 \\
%     %\bottomrule
%     \hline
%     \end{tabular}
%     \end{center}
%   \caption{Public/private scores}
% \end{table}

The main goal of our experiments is to compare metrics of models trained separately for each task and model trained on all tasks at once (Fusion). We also would like to test the assumption that the combination of similar tasks (in our case, image + text tasks: ZsOD + VQA, and tasks with an image part: HTR + ZsOD + VQA) is the most beneficial for them. 

For C2C task, we use AVATAR dataset \cite{avatar}. While the authors utilize at most 5 accepted solutions for each problem from AtCoder, Code Jam and Codeforces, we raise this number to 7 in order to increase the training dataset. For HTR, samples from IAM Handwritten Database \cite{iam} are used. For image-and-text tasks (ZsOD, VQA), we experiment with Visual Genome dataset \cite{visualgenome}; for VQA we also add ``yes/no'' questions from VQA v2 dataset \cite{vqav2}. For testing, we use English-language subsets of the datasets described in \ref{sec:tasks}, in order to compare the results with those produced by state-of-the-art single-task models.  

\begin{table}[H]
\centering
\small
\begin{tabularx}{0.47\textwidth}{@{\extracolsep{\fill}}lccccc}
\toprule[1.2pt]
\addlinespace[0.5em]
\shortstack{} & \shortstack{\textbf{C2C}} &  \shortstack{\textbf{HTR}} & \shortstack{\textbf{ZsOD}} & \shortstack{\textbf{VQA}} \\
\midrule
\addlinespace[0.5em]
{\# of samples} & 92,307 & 139,917 & 3,220,243 & 1,663,852 \\
\addlinespace[0.5em]
%\midrule
\bottomrule[1.2pt]
\addlinespace[0.5em]
\end{tabularx}
%\addlinespace[0.5em]
\caption{Number of training samples for different subtasks
\label{tab:trainingdata}}
\end{table}

In Single-task mode, all tasks are trained until the loss reached the plateau, except for ZSoD, since it requires significantly more time for convergence. In Fusion experiments, WeightBalanceSampler is used to avoid unbalanced learning. The sampler weights (see Table  \ref{tab:samplerweights}) are selected based on Single-task training so that in Fusion mode the data for each of the tasks is passed through the model as many times as in Single mode. AdamW optimizer and OneCycleLR scheduler are used for optimization. The following parameters are equal for all experiments (single and fusion tasks): warmup 0.1, pct\_start 0.1, max lr 1e-3, start\_lr=8e-6, weight decay 1e-2, beta coefficients (0.9,0.999), final\_div\_factor = 1000, 8xA100 80Gb GPUs.

The results of our experiments are introduced in Table~\ref{tab:privatescores}. Total score is the sum of scores for four subtasks, with the exception for CodeBLEU metric which is multiplied by 0.01 (refer to the section~\ref{sec:overall} for more details.). An interesting observation is that Fusion experiment exposed less over-fitting problems.

% For optimization we use AdamW optimizer and OneCycleLR scheduler with  pct\_start $= 0.1$ and final\_div\_factor $= 500$. Other parameters for scheduler were individual for all type of tasks.

\begin{table}[H]
\centering
\small
\begin{tabularx}{0.47\textwidth}{@{\extracolsep{\fill}}lccccc}
\toprule[1.2pt]
\addlinespace[0.5em]
\shortstack{training \\ setup} & \shortstack{\textbf{C2C} \\ \textbf{CodeBLEU}} &  \shortstack{\textbf{HTR} \\ \textbf{Acc}} & \shortstack{\textbf{ZsOD} \\ \textbf{F1}} & \shortstack{\textbf{VQA} \\ \textbf{Acc}} &
\textbf{Overall} \\
\midrule
\addlinespace[0.5em]
Single-task & 0.123 & 0.533 & 0.193 & 0.307 & 1.156\\
\addlinespace[0.5em]
\hline
\addlinespace[0.5em]
ZsOD + VQA & -- & -- & \textbf{0.196} & 0.313 & --\\
\addlinespace[0.5em]
\hline
\addlinespace[0.5em]
HTR + ZsOD + VQA & -- & 0.566 & \textbf{0.196} & 0.325 & --\\
\addlinespace[0.5em]
\hline
\addlinespace[0.5em]
\shortstack{Fusion} & \textbf{0.132} & \textbf{0.587}  & 0.191 & \textbf{0.327} & \textbf{1.237}\\
\addlinespace[0.5em]
%\midrule
\bottomrule[1.2pt]
\addlinespace[0.5em]
\end{tabularx}
%\addlinespace[0.5em]
\caption{Private scores for different training strategies \label{tab:privatescores}}
\end{table}

We also measured the performance of state-of-the-art single-task models (\cite{plbart, easter2, mdetr}) for each of the subtasks on our private test sets (see \ref{tab:sota}). It should be noted that the vast majority of models (including the state-of-the-art one) solve VQA task as a classification problem, which is much easier than generation (the case of our model), but at the same time such a design of a problem is far from real application. Although SOTA single-task models show higher scores for each of the subtasks (especially for VQA task, for the reasons stated earlier), the results of our ``fusion-brain'' model are rather promising, considering the versatility and simplicity of the approach, and proving the need for further research. 

\begin{table}[H]
\centering
\small
\begin{tabularx}{0.47\textwidth}{@{\extracolsep{\fill}}lccccc}
\toprule[1.2pt]
\addlinespace[0.5em]
\shortstack{} & \shortstack{\textbf{C2C} \\ \textbf{CodeBLEU} \\ \vphantom{foo} \\ PLBART} &  \shortstack{\textbf{HTR} \\ \textbf{Acc} \\ \vphantom{foo} \\ Easter2} & \shortstack{\textbf{ZsOD} \\ \textbf{F1} \\ \vphantom{foo} \\ MDETR} & \shortstack{\textbf{VQA} \\ \textbf{Acc} \\ \vphantom{foo} \\ (classification)} \\
\midrule
\addlinespace[0.5em]
score & 0.309 & 0.761 & 0.359 & 0.955 \\
\addlinespace[0.5em]
%\midrule
\bottomrule[1.2pt]
\addlinespace[0.5em]
\end{tabularx}
%\addlinespace[0.5em]
\caption{Scores of SOTA models on private test sets \label{tab:sota}}
\end{table}

% \begin{table}[H]
% \centering
% \small
% \begin{tabularx}{0.47\textwidth}{@{\extracolsep{\fill}}lccccc}
% \toprule[1.2pt]
% \addlinespace[0.5em]
% \shortstack{training \\ setup} & \shortstack{\textbf{C2C} \\ \textbf{CodeBLEU}} &  \shortstack{\textbf{HTR} \\ \textbf{Acc}} & \shortstack{\textbf{ZsOD} \\ \textbf{F1}} & \shortstack{\textbf{VQA} \\ \textbf{Acc}} &
% \textbf{Overall} \\
% \midrule
% \addlinespace[0.5em]
% Single-task & 0.34 & \textbf{0.63} & 0.17 & 0.25 & 1.39\\
% \addlinespace[0.5em]
% \hline
% \addlinespace[0.7em]
% \shortstack{Fusion} & \textbf{0.39} & 0.61  & \textbf{0.21} & \textbf{0.30} & \textbf{1.51}\\
% \addlinespace[0.5em]
% %\midrule
% \bottomrule[1.2pt]
% \end{tabularx}
% \caption{Private scores for different training strategies \label{tab:privatescores}}
% \end{table}

%\begin{table}[H]
%\centering
%\small
%\begin{tabularx}{0.47\textwidth}{@{\extracolsep{\fill}}lcccc}
%\toprule[1.2pt]
%\addlinespace[0.5em]
%\shortstack{training \\ setup} & \shortstack{\textbf{Training} \\ \textbf{time (hours)}} &  \shortstack{\textbf{Training} \\ \textbf{params}} & \textbf{CO2} \\
%\midrule
%\addlinespace[0.5em]
%Single-task & 215:00 &  3 283 978 882  & 1\\
%\addlinespace[0.5em]
%\hline
%\addlinespace[0.7em]
%\shortstack{Fusion} & 125:30 & 988 272 474  & 1 \\
%\addlinespace[0.5em]
%\midrule
%\bottomrule[1.2pt]
%\end{tabularx}
%\caption{Total parameters summarized for all 4 tasks}
%\end{table}

\subsection{Emissions reduction}

Recently, reporting energy and carbon metrics of training deep learning models has become common practice to promote energy-efficient research \cite{reportco2henderson, carbonpatterson}. In \cite{mlco2} the Machine Learning Emissions Calculator (ML CO2) is proposed, which estimates carbon emissions based on GPU type, hours spent on training, cloud provider, and region. This approach is very useful as it does not require reproducing the training process \cite{aigambit}. According to ML CO2, we estimate (see Table~\ref{tab:co2}) that training the model in the fusion setup generates almost one third less CO2eq (carbon-dioxide equivalent) than when training in a single-task regime, thus proving multi-task learning to be more energy-efficient and climate-friendly. More detailed information on parameters, including training time, for each training setup can be find in \ref{tab:trainingparams}.

\begin{table}[H]
\centering
\small
\begin{tabularx}{0.47\textwidth}{@{\extracolsep{\fill}}lccc}
\toprule[1.2pt]
\addlinespace[0.5em]
\shortstack{training \\ setup} & \shortstack{\textbf{Training} \\ \textbf{time (hours)}} &  \shortstack{\textbf{Training} \\ \textbf{params}} & \textbf{CO2 (kg)}\\
\midrule
\addlinespace[0.5em]
Single-task & 48.5 &  3,283,978,882 & 59.20\\
\addlinespace[0.5em]
\hline
\addlinespace[0.7em]
\shortstack{Fusion} & \textbf{35} & \textbf{988,272,474} & \textbf{42.72}\\
\addlinespace[0.5em]
%\midrule
\bottomrule[1.2pt]
\addlinespace[0.5em]
\end{tabularx}
%\addlinespace[0.5em]
\caption{Total parameters summarized for all 4 tasks 
\label{tab:co2}}
\end{table}

% in Table~\ref{tab:co2}

%\section{Evaluation Metrics} %\label{sec:metrics}

% \subsection{F1-score for ZsOD}
% To assess the quality of the detection model we use an F1-score. 
% In our non-trivial case of multi-label object detection we calculate these statistics as follows:
% \begin{itemize} 
%     \item FN -- for a given label the model has not predicted or predicted not all required bounding boxes;
%     \item TP -- a bounding box predicted by the model has IoU-score (intersection-over-union) with at least one of the ground truth bounding boxes for considered label higher than 0.5;
%     \item FP -- a predicted bounding box has IoU score less than 0.5 with all ground truth bounding boxes or there is no object of the given label on the image, yet model has predicted boundaries for it instead of returning empty list.
% \end{itemize}

%\subsection{Overall metric}
%Overall metric is the sum of scores for four subtasks. Since all tasks are scored from 0 to 1 (the only exception is the CodeBLEU metric: it may take values within the range from 0 to 100 -- with a view to normalize it, the metric is multiplied by 0.01), final result can range from 0 to 4.

\section{Conclusion}

In this paper we have presented the AI Journey 2021 Challenge called Fusion Brain \cite{fbchallenge} -- to the best of our knowledge, a first competition that is dedicated to the creation of the unified architecture which could deal with different modalities and solve 4 tasks for vision, language and programming code: Code2code Translation, Handwritten Text recognition, Zero-shot Object Detection, and Visual Question Answering. To test the participants' submissions, the datasets for each task were created; we also have described how the data were prepared. To date, the Russian part of the proposed dataset for HTR task is the largest Russian handwritten dataset in the world. We also came up with a task statement and created a leaderboard for the Fusion Brain Challenge. Actually, there were 41 teams that took part in the competition and made at least one submission, and 513 submissions in total (refer to \cite{dsworks} and section \ref{sec:private}). Moreover, according to our estimations, the proposed multitask fusion approach proves to be more energy-efficient and therefore provides a CO2 emissions reduction.

%%
%% The acknowledgments section is defined using the "acks" environment
%% (and NOT an unnumbered section). This ensures the proper
%% identification of the section in the article metadata, and the
%% consistent spelling of the heading.

\section{Acknowledgments}
We would like to thank Sber and SberCloud for granting the GPU-resources to experiment with different architectures and for supporting the Fusion Brain Challenge. 
\vspace{5mm}

\textit{The article has been recently submitted for review to the "Computer Optics" journal.}

\bibliographystyle{plain}
\bibliography{ml}

\section{Appendix}

\subsection{F1-score for ZsOD}
\label{sec:f1-zsod}
To assess the quality of the detection model we use an F1-score:
\[
F1= 2 \cdot \frac{\text{Recall} \cdot \text{Precision}}{\text{Recall} + \text{Precision}}
\]
The F1-score is calculated based on Recall and Precision, which, in turn, depend on a set of prediction statistics - true positive (TP), false positive (FP) and false negative (FN):
\[
Recall= \frac{\text{True\ Positive}}{\text{True\ Positive} + \text{False\ Negative}},
\]

\[
Precision= \frac{\text{True\ Positive}}{\text{True\ Positive} + \text{False\ Positive}}.
\]
In our non-trivial case of multi-label object detection we calculate these statistics as follows:
\begin{itemize} 
    \item FN -- for a given label the model has not predicted or predicted not all required bounding boxes;
    \item TP -- a bounding box predicted by the model has IoU-score (intersection-over-union) with at least one of the ground truth bounding boxes for considered label higher than 0.5;
    \item FP -- a predicted bounding box has IoU score less than 0.5 with all ground truth bounding boxes or there is no object of the given label on the image, yet model has predicted boundaries for it instead of returning empty list.
\end{itemize}
\subsection{Overall metric}
\label{sec:overall}
Total score is the sum of scores for four subtasks. Since all tasks are scored from 0 to 1 (the only exception is the CodeBLEU metric: it may take values within the range from 0 to 100 -- with a view to normalize it, the metric is multiplied by 0.01), final result can range from 0 to 4.

\subsection{Sampler weights}

\begin{table}[H]
\centering
\small
\begin{tabularx}{0.47\textwidth}{@{\extracolsep{\fill}}lccccc}
\toprule[1.2pt]
\addlinespace[0.5em]
\shortstack{training \\ setup} & \shortstack{\textbf{C2C}} &  \shortstack{\textbf{HTR}} & \shortstack{\textbf{ZsOD}} & \shortstack{\textbf{VQA}} \\
\midrule
\addlinespace[0.5em]
ZsOD + VQA & -- & -- & 0.78 & 0.22 \\
\addlinespace[0.5em]
\hline
\addlinespace[0.5em]
HTR + ZsOD + VQA & -- & 0.19  & 0.64 & 0.17 \\
\addlinespace[0.5em]
\hline
\addlinespace[0.5em]
Fusion & 0.04 & 0.18  & 0.61 & 0.17 \\
\addlinespace[0.5em]
\bottomrule[1.2pt]
\addlinespace[0.5em]
\end{tabularx}
%\addlinespace[0.5em]
\caption{Weights of WeightBalanceSampler for different tasks \label{tab:samplerweights}}
\end{table}

\begin{figure*}[ht!]
%   \centering
  \includegraphics[width=1.\textwidth]{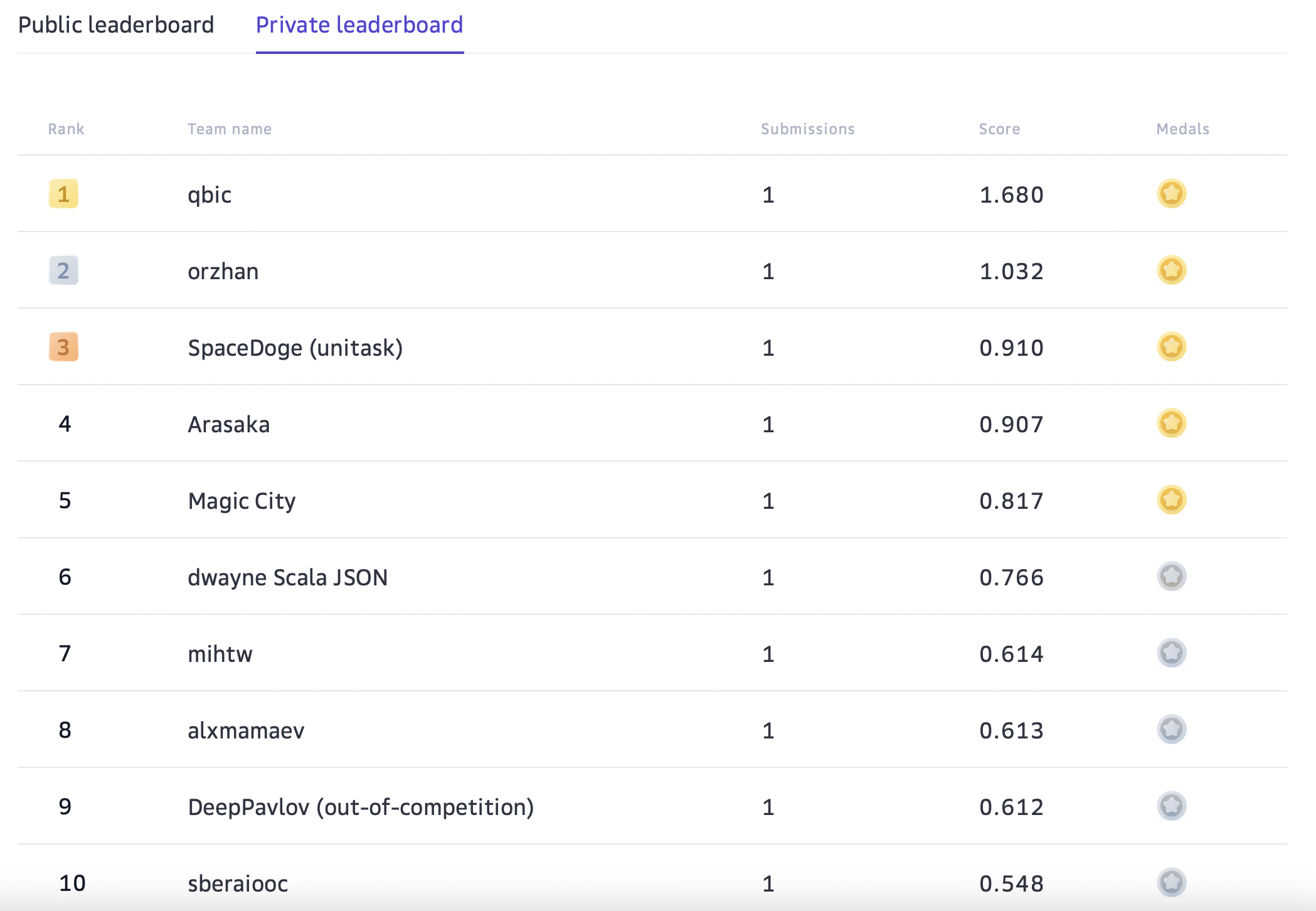}
  \caption{Top-10 participants sorted by Total score in private leaderboard}
  \label{privateleaderboardfirst}
\end{figure*}

\subsection{Training parameters}

\begin{table}[H]
\centering
\small
\begin{tabularx}{0.47\textwidth}{@{\extracolsep{\fill}}lccccc}
\toprule[1.2pt]
\addlinespace[0.5em]
\shortstack{training \\ setup} & \shortstack{\textbf{batch} \\ \textbf{size}} &  \shortstack{\textbf{\# of} \\ \textbf{steps}} & \shortstack{\textbf{time} \\ \textbf{(hrs)}} & \shortstack{\textbf{CO2} \\ \textbf{(kg)}} \\
\midrule
\addlinespace[0.5em]
C2C & 8 & 69,264 & 8.5 & 10.4 \\
\addlinespace[0.5em]
\hline
\addlinespace[0.5em]
HTR & 32 & 70,305 & 4.5 & 5.52 \\
\addlinespace[0.5em]
\hline
\addlinespace[0.5em]
ZsOD & 64 & 119,510 & 28.5 & 34.8 \\
\addlinespace[0.5em]
\hline
\addlinespace[0.5em]
VQA & 64 & 32,500 & 7 & 8.56 \\
\addlinespace[0.5em]
\hline
\addlinespace[0.5em]
ZsOD + VQA & 64 & 175,000 & 32 & 39.04 \\
\addlinespace[0.5em]
\hline
\addlinespace[0.5em]
HTR + ZsOD + VQA & 64 & 180,000 & 34 & 41.44 \\
\addlinespace[0.5em]
\hline
\addlinespace[0.5em]
\shortstack{Fusion} & 96 & 140,000 & 35 & 42.72 \\
\addlinespace[0.5em]
%\midrule
\bottomrule[1.2pt]
\addlinespace[0.5em]
\end{tabularx}
%\addlinespace[0.5em]
\caption{Several training parameters for different training strategies \label{tab:trainingparams}}
\end{table}

\subsection{Private leaderboard}
\label{sec:private}
We provide the private leaderboard on the Fusion Brain Challenge (see Figure \ref{privateleaderboardfirst}). Metrics of the winner of the competition are the following:
\begin{table}[H]
\centering
\small
\begin{tabularx}{0.47\textwidth}{@{\extracolsep{\fill}}lccccc}
\toprule[1.2pt]
\addlinespace[0.5em]
\shortstack{team \\ name} & \shortstack{\textbf{C2C} \\ \textbf{CodeBLEU}} &  \shortstack{\textbf{HTR} \\ \textbf{Acc}} & \shortstack{\textbf{ZsOD} \\ \textbf{F1}} & \shortstack{\textbf{VQA} \\ \textbf{Acc}} &
\textbf{Total} \\
\midrule
\addlinespace[0.5em]
qbic & 0.320 & 0.744 & 0.250 & 0.365 & 1.680\\
\addlinespace[0.5em]
\hline
\addlinespace[0.5em]
orzhan & 0.233 & 0.314  & 0.166 & 0.318 & 1.032\\
\addlinespace[0.5em]
\hline
\addlinespace[0.5em]
Arasaka & 0.218 & 0.377 & 0.074 & 0.237 & 0.907\\
\addlinespace[0.5em]
%\hline
%\midrule
\bottomrule[1.2pt]
\addlinespace[0.5em]
\end{tabularx}
%\addlinespace[0.5em]
\caption{Top-3 private scores of the multi-modal and multi-task models provided by participants of the Fusion Brain Challenge \label{tab:privateleaderboardfirst}}
\end{table}

\end{document}